\documentclass[journal,twoside,web]{ieeecolor}
\usepackage{tmi}
\usepackage{cite}
\usepackage{amsmath,amssymb,amsfonts}
\usepackage{graphicx}
\usepackage{textcomp}
\usepackage{bm}
\usepackage{soul}
\usepackage{multirow}
\usepackage{subfig}
\usepackage{algorithm}
\usepackage{algpseudocode}

\usepackage[pagebackref=false,breaklinks=true,letterpaper=true,colorlinks,bookmarks=true]{hyperref}

\def\BibTeX{{\rm B\kern-.05em{\sc i\kern-.025em b}\kern-.08em
	T\kern-.1667em\lower.7ex\hbox{E}\kern-.125emX}}
\def\Fig#1{{Fig.~\ref{fig:#1}}}
\def\Eq#1{{Eq.~\eqref{eq:#1}}}
\def\Table#1{{Table~\ref{tbl:#1}}}
\def\Sec#1{{Sec.~\ref{sec:#1}}}

\def\npos{n_\text{pos}}
\def\nneg{n_\text{neg}}
\def\nlcand{n^l_\text{cand}}
\def\ngrand{n^g_\text{rand}}
\def\FtMap{F}
\def\FtVec{\bm{f}}
\def\FtVecNeg{\bm{h}}

\begin{document}
\title{SAM: Self-supervised Learning of Pixel-wise Anatomical Embeddings in Radiological Images}
\author{Ke Yan*, Jinzheng Cai, \IEEEmembership{Member, IEEE}, Dakai Jin, \IEEEmembership{Member, IEEE}, Shun Miao, Dazhou Guo, \\Adam P.~Harrison, Youbao Tang, Jing Xiao, Jingjing Lu*, Le Lu, \IEEEmembership{Fellow, IEEE}
	\thanks{K.~Yan is with Alibaba Group DAMO Academy, Beijing, 100026, China; D.~Jin, D.~Guo, and L.~Lu are with Alibaba Group DAMO Academy, New York, NY 10014, USA.}
	\thanks{J.~Cai, S.~Miao, A.~P.~Harrison and Y.~Tang are with PAII Inc., Palo Alto, CA 94306, USA.}
	\thanks{J. Xiao is with Ping An Insurance (Group) Company of China, Ltd., Shenzhen, 510852, China.} 
	\thanks{J.~Lu is with Beijing United Family Hospital, Beijing, 100015, China.}
	\thanks{Corresponding authors: Ke Yan (email: yankethu@gmail.com); Jingjing Lu (email: cjr.lujingjing@vip.163.com)}}

\maketitle

\begin{abstract}
	Radiological images such as computed tomography (CT) and X-rays render anatomy with intrinsic structures. Being able to reliably locate the same anatomical structure across varying images is a fundamental task in medical image analysis. In principle it is possible to use landmark detection or semantic segmentation for this task, but to work well these require large numbers of labeled data for each anatomical structure and sub-structure of interest. A more universal approach would learn the intrinsic structure from \emph{unlabeled} images. We introduce such an approach, called Self-supervised Anatomical eMbedding (SAM). SAM generates semantic embeddings for each image pixel that describes its anatomical location or body part. To produce such embeddings, we propose a pixel-level contrastive learning framework. A coarse-to-fine strategy ensures both global and local anatomical information are encoded. Negative sample selection strategies are designed to enhance the embedding's discriminability. Using SAM, one can label any point of interest on a template image and then locate the same body part in other images by simple nearest neighbor searching. We demonstrate the effectiveness of SAM in multiple tasks with 2D and 3D image modalities. On a chest CT dataset with 19 landmarks, SAM outperforms widely-used registration algorithms while only taking 0.23 seconds for inference. On two X-ray datasets, SAM, with only one labeled template image, surpasses supervised methods trained on 50 labeled images. We also apply SAM on whole-body follow-up lesion matching in CT and obtain an accuracy of 91\%. SAM can also be applied for improving image registration and initializing CNN weights.
\end{abstract}


\section{Introduction}

\begin{figure}[t]
	\centerline{\includegraphics[width=\columnwidth,trim=0 250 740 0, clip]{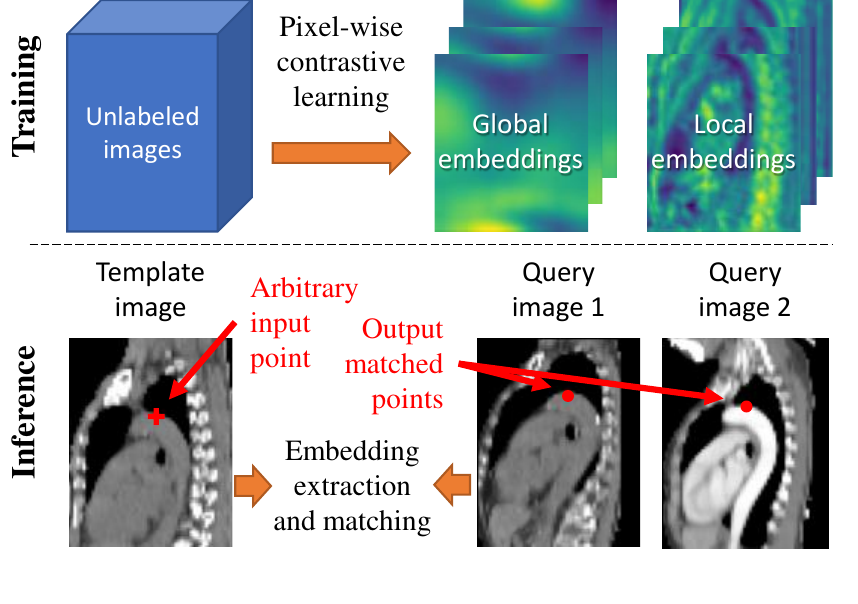}} 
	\caption{Self-supervised anatomical embedding (SAM) and its application of anatomical location matching.}
	\label{fig:overview} 
\end{figure}

In medical image analysis, many tasks essentially {aim} at locating certain anatomical structures. For example, {landmark detection is} useful for diagnosis, quantification, therapy evaluation and planning, or as a preprocessing step for other downstream tasks~\cite{Litjens2017survey, Zheng2015landmark}. Lesion matching locates follow-up lesions in multiple images, which is an important clinical task for radiologists to longitudinally track disease progress~\cite{Cai2021DLT}.
However, supervised algorithms~~\cite{Zheng2015landmark, Cai2021DLT} need sufficient labeled data for training. Annotating medical images is laborious, expensive, and requires considerable expertise~\cite{Tajbakhsh2020imperfect}. Moreover, supervised models are restricted to only discovering landmarks or lesions that were \emph{a priori} labeled and trained, but many labeled datasets are limited in their coverage and fine-grainedness. 

Human organs are intrinsically structured, so there is an inherent consistency underlying their appearance and layout in radiological images such as computed tomography (CT) and X-rays. Intuitively, we can design an algorithm to discover these intrinsic patterns by learning from \emph{unlabeled} images. Given a certain anatomical location (e.g., top of aortic arch) in one image, the trained algorithm should be able to find the corresponding location in other images (see \Fig{overview}), although it 
was not trained specifically to detect such landmarks. We aim to develop a universal algorithm that learns from unlabeled radiological images to detect arbitrary points of interest. It can generate embeddings on each image pixel to encode its anatomical context information, so that the same body part in different images express similar embeddings and can be retrieved by simple nearest neighbor searching. In this paper, we use the terms anatomical location or body part to describe the semantic location of a pixel, such as ``center of the right lobe of thyroid'' or ``femur head''.

Learning anatomical embeddings that are universal across body parts, without anatomical labels in training, is a challenging task. For example, in CT images the algorithm has to memorize the 3D contextual appearance of numerous body parts so as to generate globally distinguishable embeddings. Meanwhile, it needs to encode local information to differentiate adjacent structures with similar appearance for accurate localization. In addition, the embeddings should be robust to the size, shape, intensity, and texture diversity of body parts caused by inter-subject variation, organ deformation, contrast injection, and pathological changes.

In this work, we propose Self-supervised Anatomical eMbedding (SAM) to achieve these goals. We adopt the recent self-supervised contrastive learning~\cite{Wu2018instdis, Chen2020SimCLR, He2020MoCo} approach 
and propose a pixel-level contrastive learning framework to differentiate pixels of varying body parts. 
During training, we use data augmentation to create synthetic body part pairs and simulate appearance changes across different images. 
To cover both global and local information, we design a coarse-to-fine architecture with two-level embeddings. The global embedding is trained to distinguish every body part on a coarse scale, helping the local embedding to focus on a smaller region to differentiate with finer features. 
The selection of negative samples (in our case, dissimilar body parts) is a crucial step of contrastive learning. We design several strategies to carefully control the difficulty and diversity of the selected negative samples. Finally, during inference, embeddings can be matched efficiently using 2D or 3D convolutional operations.

Due to its unsupervised and universal nature, SAM can be trained easily and applied to various tasks. A natural application is  ``one-shot'' landmark detection with one labeled template image
. We conduct such experiments on a chest CT dataset and two X-ray datasets (pelvis and hand)~\cite{Li2020landmark}. On the chest CT dataset, SAM outperforms widely-used deformable registration algorithms~\cite{Avants2008ANTS, heinrich2013mrf,  Rueckert1999FFD, Balakrishnan2019VoxelMorph} while only taking 0.23 seconds to infer an image. On the X-ray datasets, our method with one labeled template outperforms the supervised HRNet~\cite{Sun2019HRNet} trained on 50 labeled images
. Another application is lesion matching~\cite{Yan2018graph, Rafael2021matching}. On 3,556 lesion pairs manually annotated in the DeepLesion dataset~\cite{Yan2018DeepLesion, Yan2018graph}, SAM outperforms 
previous supervised matching algorithms~\cite{Yan2018graph, Yan2019Lesa, Cai2021DLT} (91\% versus 82\% in accuracy). 
SAM can also be applied to improve unsupervised deformable image registration by providing corresponding points between images, semantic features, and losses. Our SAM-enhanced registration algorithm~\cite{Liu2021SAME} achieves superior accuracy compared to existing learning-based~\cite{Balakrishnan2019VoxelMorph} or optimization-based methods~\cite{Rueckert1999FFD, Avants2008ANTS, heinrich2013mrf}.

Our major contributions are three-fold: {\bf 1)} The problem of universal anatomical embedding learning is tackled for the first time. We demonstrate that discriminative and robust embeddings can be learned in a self-supervised manner. {\bf 2)} A pixel-level contrastive learning framework is proposed with a coarse-to-fine architecture and customized negative sampling strategies. 
The proposed algorithm is easy to train and fast. 
{\bf 3)} Most importantly, we show the learned embeddings are widely applicable in various applications (landmark detection, lesion matching, registration, etc.) on different image modalities (3D CT and 2D X-ray) of varying body parts (chest, hand, pelvis, etc.). Similar or superior performance is achieved compared to existing registration or supervised algorithms.

\section{Related Work}

{\bf Self-supervised learning (SSL).} In SSL, visual features are learned from unlabeled data by optimizing a heuristic pretext task, such as predicting image rotations~\cite{Gidaris2018rot} and inpainting~\cite{Pathak2016inpainting}. Contrastive learning algorithms~\cite{Wu2018instdis, Chen2020SimCLR, He2020MoCo} use a simple paradigm of data augmentation and image instance discrimination and achieve impressive performance. Recently, some SSL methods were developed to learn dense representations, 
e.g., VADeR~\cite{Pinheiro2020dense} and DenseCL~\cite{Wang2021Dense}. 
They focus on network pretraining and need finetuning on downstream tasks such as object detection and segmentation. However, our goal is learning anatomical embeddings, which is coherent in both training and inference; thus, no finetuning is needed. Both VADeR and DenseCL are designed for natural images and rely on image-wise contrast as initialization~\cite{Pinheiro2020dense} or as an additional loss~\cite{Wang2021Dense}, which does not apply to radiological images. This is because each radiological image contains the same set of body parts, so it is hard to define image-wise negative samples as all images are similar. Our method is specially designed for radiological images with strategies crucial to make the embeddings effectively discriminative and fine-grained, i.e., a coarse-to-fine structure and negative sampling algorithms.

\begin{figure*}[ht!]
	\centering
	\includegraphics[width=1.0\textwidth,trim=0 300 220 0, clip]{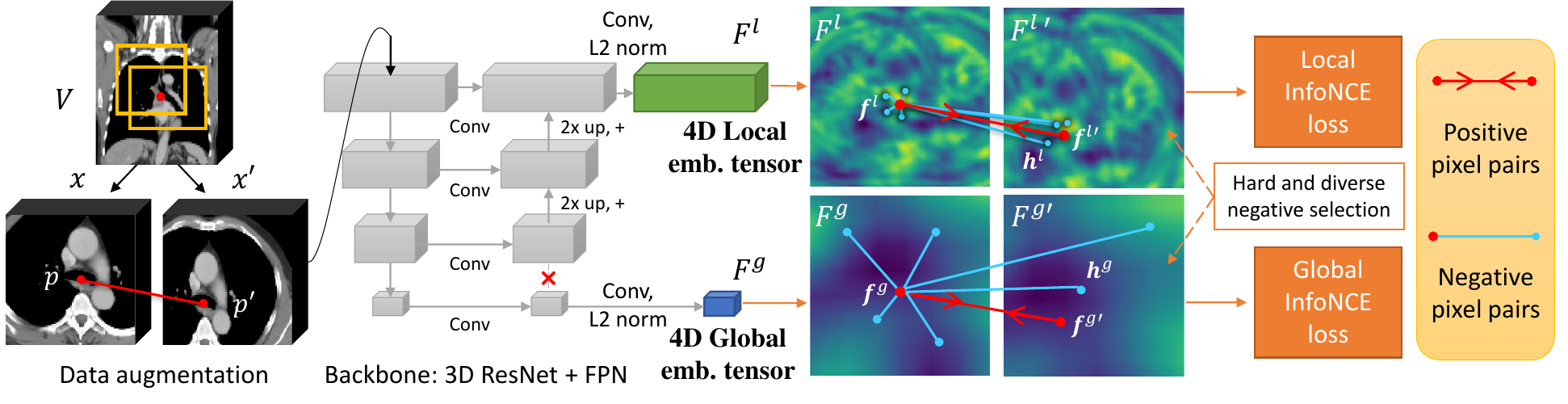} 
	\caption{The learning process of SAM. We first do random data augmentation to an unlabeled image, then send the two transformed patches to a network to generate a global and a local embedding vector for each pixel (\Sec{mtd:network}). One embedding channel of an axial image slice is visualized in the figure with the global embedding zoomed in. Next, we use a global and a local InfoNCE loss~\cite{VanDenOord2018CPC} to encourage matched pixels (positive pairs) to have similar embeddings, while push unmatched ones (negative pixels) apart (\Sec{mtd:pair_smp}). The negative pixels are random pixels in the two patches guided by hard and diverse sample selection strategies (\Sec{mtd:neg_smp}).
	} 
	\label{fig:training} 
\end{figure*}

Radiological images are suitable for SSL because of the intrinsic structure of the human anatomy. Related studies can be classified into three major categories. Content restoration algorithms~\cite{Zhou2020Genesis, Feng2020Parts, Chen2019restore} reconstruct original images from corrupted or cropped ones, aiming to pretrain network weights. For instance, Models Genesis~\cite{Zhou2020Genesis} designs a series of image corruption strategies. Its pretrained models outperforms randomly initialized ones in a wide range of downstream applications covering both segmentation and classification. Transformation prediction algorithms first do data augmentation to the image, and then predict the augmentation parameters, such as translation~\cite{Blendowski2019ssl}, patch order~\cite{Zhuang2019rubik}, and rotation~\cite{Tajbakhsh2019surrogate}. An example is Rubik's cube~\cite{Zhuang2019rubik}, which uses cube rearrangement and cube rotation prediction as proxy tasks to enforce networks to learn translational and rotational invariant features from raw 3D data. Image-wise~\cite{Azizi2021big}, patch-wise~\cite{Chaitanya2020contrastive} and superpixel-wise~\cite{Ouyang2020superpx} contrastive learning have also been proposed to pretrain networks~\cite{Azizi2021big} or learn representations for few-shot image segmentation~\cite{Chaitanya2020contrastive, Ouyang2020superpx}. 
None of these works are designed for universal and fine-grained anatomical point matching. In this task, each pixel must be encoded to be distinguishable from other body parts, which is SAM's explicit training objective. Other self-supervised algorithms use different proxy tasks that are not directly applicable in this task.
Additionally, methods in \cite{Chaitanya2020contrastive} and \cite{Ouyang2020superpx} require relatively complex multi-stage training or superpixel extraction, while our method is simple and end-to-end trainable.

{\bf Anatomy-related learning.} Anatomical landmark detection can be used to initialize or aid many medical imaging tasks~\cite{Payer2019landmark, Li2020landmark, Chen2020fracture, Ghesu2019RL, Vlontzos2019RL}. Heatmap and coordinate-based methods have been studied~\cite{Li2020landmark}. Deep adaptive graph was proposed in~\cite{Li2020landmark} to capture relationships among landmarks and achieved state-of-the-art results in three X-ray datasets. However, these methods all need training labels and can only predict the predefined anatomical points, whereas our method is unsupervised and uses embedding matching to find arbitrary landmarks of interests. Anatomy-specific classification~\cite{Roth2015anatomy} or regression~\cite{Yan2018UBR} methods can predict predefined body part classes or scores of CT slices. Compared to their slice-wise coarse predictions, our pixel-wise embedding is more fine-grained.

{\bf Image registration} is popularly used to find pixel-wise correspondences in medical images. Traditional algorithms~\cite{Rueckert1999FFD, heinrich2013mrf, Avants2008ANTS} iteratively optimize a predefined objective function. Recent deep learning-based methods~\cite{Balakrishnan2019VoxelMorph, Sokooti2017registration} directly predict a deformation field to align an image pair using the spatial transformer network~\cite{jaderberg2015spatial}. 
They are fast in inference but often less accurate than state-of-the-art traditional algorithms~\cite{heinrich2013mrf, Liu2020JSSR}. Registration needs an image pair as input and aims to align each pixel via fundamental appearance-based similarity measures. By contrast, SAM tackles the pixel matching problem in a different perspective by semantically representing each pixel given one single image. Its core application is anatomical point matching instead of full-volume registration, but we will show it can be readily used to improve existing registration algorithms~\cite{Liu2021SAME}.

\section{Method}
\label{sec:method}

The objective of self-supervised anatomical embedding (SAM) is to encode the semantic anatomical information of each pixel, so that similar body parts in different images have similar embeddings. Inspired by previous works on contrastive learning~\cite{Wu2018instdis, Chen2020SimCLR, He2020MoCo}, we propose a pixel-level contrastive learning framework as shown in \Fig{training}. Essentially, we let the model contrast the appearance of random pixels, so that it can gradually discover distinctive patterns of each anatomical point by itself. The proposed framework comprises the following steps: stochastic data augmentation for training patch generation, coarse-to-fine CNN for pixel-wise feature embedding, and positive \& negative sample selection for contrastive loss computation. In this section, we elaborate our framework for 3D images such as CT. It is straightforward to adapt SAM to 2D images such as X-ray by changing the network from 3D to 2D.



\subsection{Coarse-to-Fine Network Architecture}
\label{sec:mtd:network}

As shown in \Fig{training}, in training, we first take an unlabeled CT volume $ V $, then crop two 3D patches with random location and size, and resize them to the same shape, namely $ x,x' \in \mathbb{R}^{d\times h\times w} $. The patches are sent to a fully-convolutional network to extract pixel-wise embeddings. To learn universal anatomical embeddings, on the one hand, SAM needs to memorize the 3D contextual appearance of each body part so its embedding is distinguishable from all other body parts. On the other hand, it needs to encode local information to differentiate adjacent structures with similar appearance for accurate localization. To get the best of both worlds, we propose a coarse-to-fine architecture that predicts a global embedding and a local one for each pixel. A lightweight 3D ResNet-18~\cite{He2016resnet} is adopted as backbone with a feature pyramid network (FPN)~\cite{Lin2016Pyramid} to fuse multi-scale features. The ResNet-18 is initialized with ImageNet pretrained weights using the inflated 3D (I3D) technique\footnote{Suppose a pretrained 2D filter is $ s\times s $ in size, I3D essentially repeats it in the $ z $-dimension $ d $ times to obtain an 3D filter of $ s\times s\times d $, and then rescales the 3D filter by dividing by $ d $.}~\cite{Carreira2017I3D}. The 4D global embedding tensor $ \FtMap^g $ and the local one $ \FtMap^l $ are both generated from the FPN features using separate $ 3\times3\times3 $ convolutions and L2 normalization layers. $ \FtMap^g $ is from the coarsest FPN level with a larger stride and more abstract features, while $ \FtMap^l $ is from the finest FPN level with a smaller stride and detailed features. Specifically, we cut the connection between the coarsest FPN feature and its upper level to make $ \FtMap^g $ and $ \FtMap^l $ more independent, see \Fig{training}. Examples of the learned $ \FtMap^g $ and $ \FtMap^l $ can be found in Figs.~\ref{fig:training} and \ref{fig:overview}, where $ \FtMap^l $ contains more high-frequency details.

\subsection{Pixel Pairs and Loss Function}
\label{sec:mtd:pair_smp}

In this step, we {select} matched and unmatched pixel pairs to train SAM. Note that although we crop two patches $ x,x' $ from the original volume $V$, we aim to contrast the \emph{pixels} inside the patches, which is different from previous works that contrast patches~\cite{Chen2020SimCLR, Blendowski2019SSL3D, Chaitanya2020contrastive}.
When the two patches overlap, we can find a pixel pair $ p\in x, p'\in x' $ from them so that $ p $ and $ p' $ actually correspond to the same pixel in $V$, see \Fig{training}. They are treated as a positive pair. 
We randomly sample $ \npos $ pixel pairs from the overlapping area. The sampled pixels do not need to be close to patch centers, because SAM needs to learn to embed all pixels in an image, including those at the center and borders. The positive embeddings at $ p_i, p'_i$ are $ \FtVec_i, \FtVec_i'\in\mathbb{R}^{c}, 1\leq i\leq\npos $. 
For each positive pair $(p_i,p'_i)$, we also find $ \nneg $ pixels as negative pixels. They are randomly sampled from $x$ and $x'$ {(including the overlapping area)} as long as their distance from $p_i$ and $p'_i$ is larger than $ \delta $ ($\delta=3\text{mm}$ for CT {and 5 pixels for X-ray}). We denote the embeddings at the negative pixels of pair $i$ as $ \FtVecNeg_{ij}\in\mathbb{R}^{c}, 1\leq j\leq\nneg $, $ j $ being the index of negative pixels. The InfoNCE loss~\cite{VanDenOord2018CPC} for pixels in this patch pair is defined as
\begin{equation}\label{eq:infoNCE}
	\mathcal{L}=-\sum_{i=1}^{\npos}\log\frac{\exp(\FtVec_i\cdot\FtVec_i'/\tau)}{\exp(\FtVec_i\cdot\FtVec_i'/\tau) + \sum_{j=1}^{\nneg}\exp(\FtVec_i\cdot\FtVecNeg_{ij}/\tau)},
\end{equation}
where $ \tau=0.5 $ is a temperature parameter and $ \boldsymbol{\cdot} $ is the inner product operation. We call $ \FtVec_i $ anchor embeddings. 
Since we want \eqref{eq:infoNCE} to be symmetric with $\FtVec_i $ and $ \FtVec_i' $~\cite{Chen2020SimCLR}, another loss term with swapped $ \FtVec_i $ and $ \FtVec_i' $ is also computed and added. \Eq{infoNCE} is applied to both global and local embeddings.

In summary, we use data augmentation to simulate appearance changes across different image volumes. Given an anchor pixel $ p\in x $, its matched pixel $ p'\in x' $ represents the same anatomical location in another (simulated) image; Its unmatched pixels are in different anatomical locations that are not physically adjacent to it. \Eq{infoNCE} pulls the matched pixels (positive pairs) together and pushes unmatched ones (negative pixels) apart in the embedding space. This strategy is simple yet effective. It can make the network learn to distinguish different body parts effectively with the help of the negative sampling strategy described in the next section.

\subsection{Hard and Diverse Negative Sampling}
\label{sec:mtd:neg_smp}
We propose several strategies to select hard and diverse negative pixels and make global and local embeddings focus on different scopes. 
The global embedding $ \FtMap^g $ is responsible for differentiating all body parts on a coarse scale. To make it more discriminative, we perform online {hard} negative selection~\cite{Novotny2018geometry} to train it. For each global anchor embedding $ \FtVec^g_i $, we compute its cosine similarity map with the global embedding tensors $ \FtMap^g$ and $\FtMap^{g\prime} $ to get similarity maps $ S^g_i$ and $S^{g\prime}_i $. Then, hard negatives $ \FtVecNeg^g_{ij} $ are selected to be the $ \nneg $ embeddings with the greatest similarity with $ \FtVec^g_i $ (excluding $ \FtVec^{g}_i $ and $ \FtVec^{g\prime}_i $ themselves)
. Additionally, we further populate the negative samples to contain more {diverse} body parts. For each $ \FtVec^g_i $, we randomly sample an additional $ \ngrand $ pixels from all patches \emph{across all images} within a training batch, except the two patches of the image that $ \FtVec_i^g $ is on. Because these pixels originate from different images it is very unlikely the sampled pixels are in the same semantic position as $ \FtVec^g_i $. 

Moving on to the local embedding, it uses the same positive pixel pairs $(p_i,p'_i)$ as the global version, but with different negative samples. For each local anchor $ \FtVec^l_i $, we first compute its local similarity maps $ S^l_i, S^{l\prime}_i $ with $ \FtMap^l, \FtMap^{l\prime} $, then upsample $ S^g_i, S^{g\prime}_i $ to be the same size as $ S^l_i $, and finally use the combined global and local similarity maps $ S^g_i + S^l_i, S^{g\prime}_i + S^{l\prime}_i $ to select {hard} negatives pixels for $ \FtVec^l_i $. Examples of $ S^l $ and upsampled $ S^g $ can be found in Figs.~\ref{fig:inference} and \ref{fig:hand_sim}. The definition of hard negative pixel is similar to that in global embeddings. The pixels with higher values on the combined similarity map are harder except the positive pixel pair themselves. When the global embedding converges, $ S^g_i $ will be high only in the neighborhood of $ p_i$, so that hard negatives can be selected mostly in the local area. During inference we use the peak of $S^g+S^l$ to find matched points, so using $S^g+S^l$ for local hard negative selection is consistent with our inference process. Meanwhile, because the local embedding tensor has a high resolution, the selected local hard negatives will mostly be adjacent pixels and correlated. To improve {diversity}, we first find the top $ \nlcand > \nneg $ hard negative candidates
, then randomly sample $ \nneg $ embeddings from them to be used in \Eq{infoNCE}. Algorithm \ref{algo:SAM} summarizes the complete training procedure of SAM.

\begin{algorithm}[]
	\small
	\caption{Self-supervised anatomical embedding (SAM)}\label{algo:SAM}
	\begin{algorithmic}[1]
		\Require Unlabeled images $ \{V\} $; batch size $ b $, pixel sampling parameters $ \npos, \nneg, \ngrand, \nlcand $.
		\Ensure Trained network.
		\State In each training batch, randomly select $ b $ images from $ \{V\} $.
		\For{each image}
		\State Run random data augmentation to get a patch pair $ (x, x') $.
		\State Compute global and local embedding tensors of the two patches, $ \FtMap^g, \FtMap^{g\prime}, \FtMap^l, \FtMap^{l\prime} $.
		\State Sample positive pixel pairs $ (p_i,p'_i) $ from the overlapping area of $ x$ and $ x' $, then extract global and local positive embeddings pairs: $ (\FtVec^g_i, \FtVec^{g\prime}_i), (\FtVec^l_i,\FtVec^{l\prime}_i), 1\leq i\leq\npos$.
		\For{each positive pair $ i $}
		\State Compute similarity maps $ S^g_i, S^{g\prime}_i, S^l_i, S^{l\prime}_i $.
		\State Find global hard negatives $ \FtVecNeg^g_{ij}, 1\leq j\leq\nneg $.
		\State Sample global random negatives $ \FtVecNeg^g_{ik}, 1\leq k\leq\ngrand $ from the other $ b-1 $ patch pairs in the training batch. 
		\State Pool the hard and random negatives to get the final global negatives. The variable $\nneg$ in Eq.~1 now becomes $\nneg+\ngrand$ for global embeddings.
		\State Find local hard negatives $ \FtVecNeg^l_{ij}, 1\leq j\leq\nlcand $. Randomly sample $ \nneg $ from them as the final local negatives. The variable $\nneg$ in Eq.~1 is unchanged for local embeddings.
		\EndFor
		\State Compute the global and local InfoNCE loss $ \mathcal{L}^g, \mathcal{L}^l $ in Eq.~1. The final loss is $ \mathcal{L}=\mathcal{L}^g+\mathcal{L}^l $.
		\EndFor
	\end{algorithmic} 
\end{algorithm} 

\subsection{Application: Anatomical Point Matching}
\label{sec:mtd:infer}

\begin{figure}[t]
	\centering
	\includegraphics[width=\columnwidth,trim=0 280 600 0, clip]{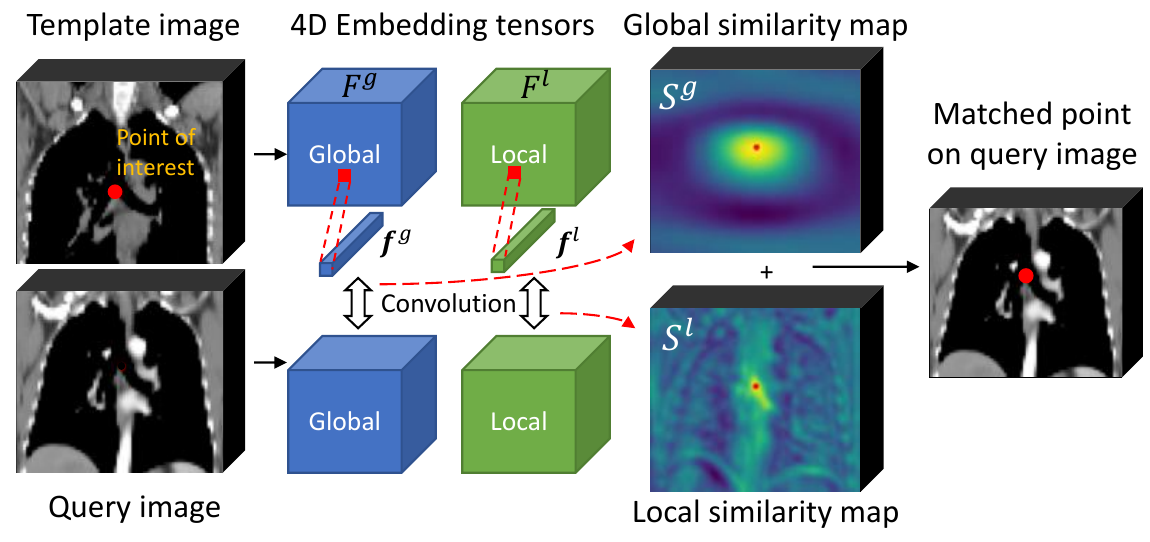} 
	\caption{The inference procedure of SAM for anatomical point matching. Red dots are the point of interest or matched point.
	} 
	\label{fig:inference} \vspace{-2mm}
\end{figure}

\Fig{inference} depicts the inference procedure of SAM for anatomical point matching. To locate a certain point of interest, we first need to label it on a template image (also known as an atlas~\cite{Wang2020LT} or a support image~\cite{Ouyang2020superpx}). Then, given an unlabeled query image, we compute the global and local embedding tensors. After extracting the anchor embedding vectors from the point of interest of the template, we compute the similarity maps $ S^g, S^l $ between the anchor vectors and the query tensors. The embeddings are L2-normalized, so the cosine similarity maps can be efficiently computed using the convolutional operation on GPU. Finally, we upsample $ S^g, S^l $ to the size of the original image, and find the peak of $ S^g+S^l$ as the matched anatomical point.

\subsection{Application: Enhancing Image Registration}
\label{sec:mtd:SAME}

SAM is designed to match sparse points. It may be possible to directly use it for full-volume registration by matching every pixel between the fixed image and the moving image, but it is highly inefficient as there are millions of pixels in a 3D CT scan. We have proposed a SAM-Enhanced deformable registration algorithm called SAME~\cite{Liu2021SAME}. It consists of three steps. (1) SAM-affine. We first sample a sparse grid on the fixed image and discard points outside the body, then use SAM to match their corresponding points on the moving image. The matches with similarity scores higher than a threshold are kept. With these point pairs, an affine transformation matrix can be estimated by simple least squares fitting. (2) SAM-coarse, which uses SAM to compute a new correspondence grid on the transformed images to interpolate a coarse-level deformation field. These two steps are efficient, require no additional training, and can provide a good initialization for the final step. (3) Lastly, SAM-VoxelMorph, which enhances the deep learning-based VoxelMorph registration method~\cite{Balakrishnan2019VoxelMorph} by incorporating SAM-based correlation features and an additional SAM-based similarity loss. More details can be found in our published work~\cite{Liu2021SAME}. SAME is unsupervised and only needs a trained SAM model.

\section{Experiments}
\label{sec:exp}

\begin{figure}[]
	\centering
	\subfloat[3D backbone for CT]{
		\label{fig:network_3d}
		\includegraphics[width=.48\columnwidth,trim=0 150 0 0, clip]{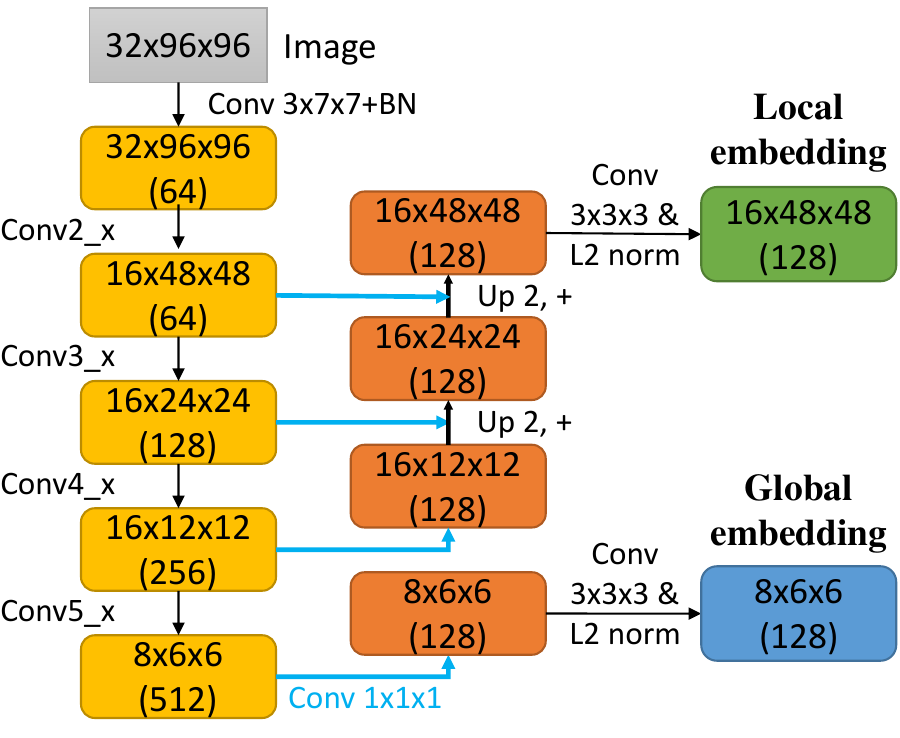}
	}
	\subfloat[2D backbone for X-ray]{
		\label{fig:network_2d}
		\includegraphics[width=.48\columnwidth,trim=0 150 0 0, clip]{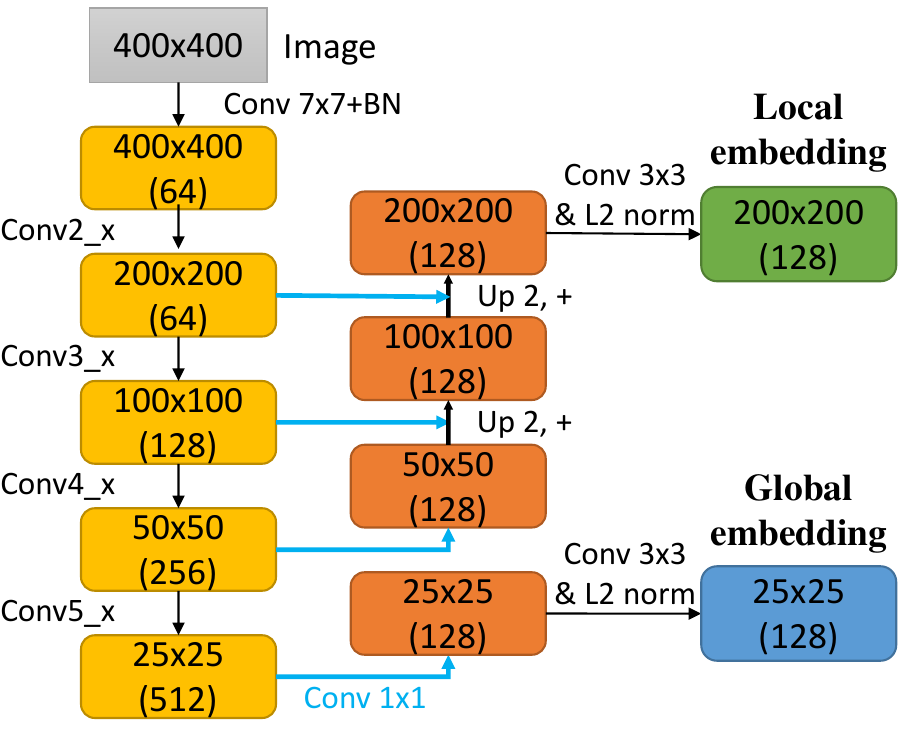}
	}
	\caption{{Detailed coarse-to-fine network structures.} The shapes of images and feature maps in training are shown in the format of $ z\times y\times x \;(channel) $. The backbones are based on (a) 3D ResNet-18 and (b) 2D ResNet-34~\cite{He2016resnet}. Downsampling is done by max pooling in Conv2\_x and by convolution with stride in other blocks. To enlarge the resolution of the feature maps, we use stride 1 in the first convolutional layer and set the $ z $ stride to 1 in Conv3\_x and Conv4\_x.}
	\label{fig:network}
	\vspace{-3mm}
\end{figure}
We evaluate SAM on four diverse tasks: 3D landmark detection on chest CT, 2D landmark detection on hand and pelvic X-rays, and 3D universal lesion matching on CT. 
Our self-supervised method is compared with widely-used registration methods~\cite{heinrich2013mrf, Balakrishnan2019VoxelMorph, Avants2008ANTS, Rueckert1999FFD}, supervised landmark detection~\cite{Payer2019landmark, Li2020landmark, Sun2019HRNet} and lesion matching~\cite{Cai2021DLT, Yan2018graph, Yan2019Lesa} methods, and other feature extraction~\cite{Calonder2010BRIEF} and network pretraining methods~\cite{Chen2019Med3D, Zhou2020Genesis}. We will briefly describe our results on 3D CT registration and direct readers to~\cite{Liu2021SAME} for more details. As extensions of SAM, we further explore using SAM for network initialization and training SAM with auxiliary label information.

\begin{figure*}[!ht]
	\centering
	\includegraphics[width=\textwidth,trim=0 100 75 0, clip]{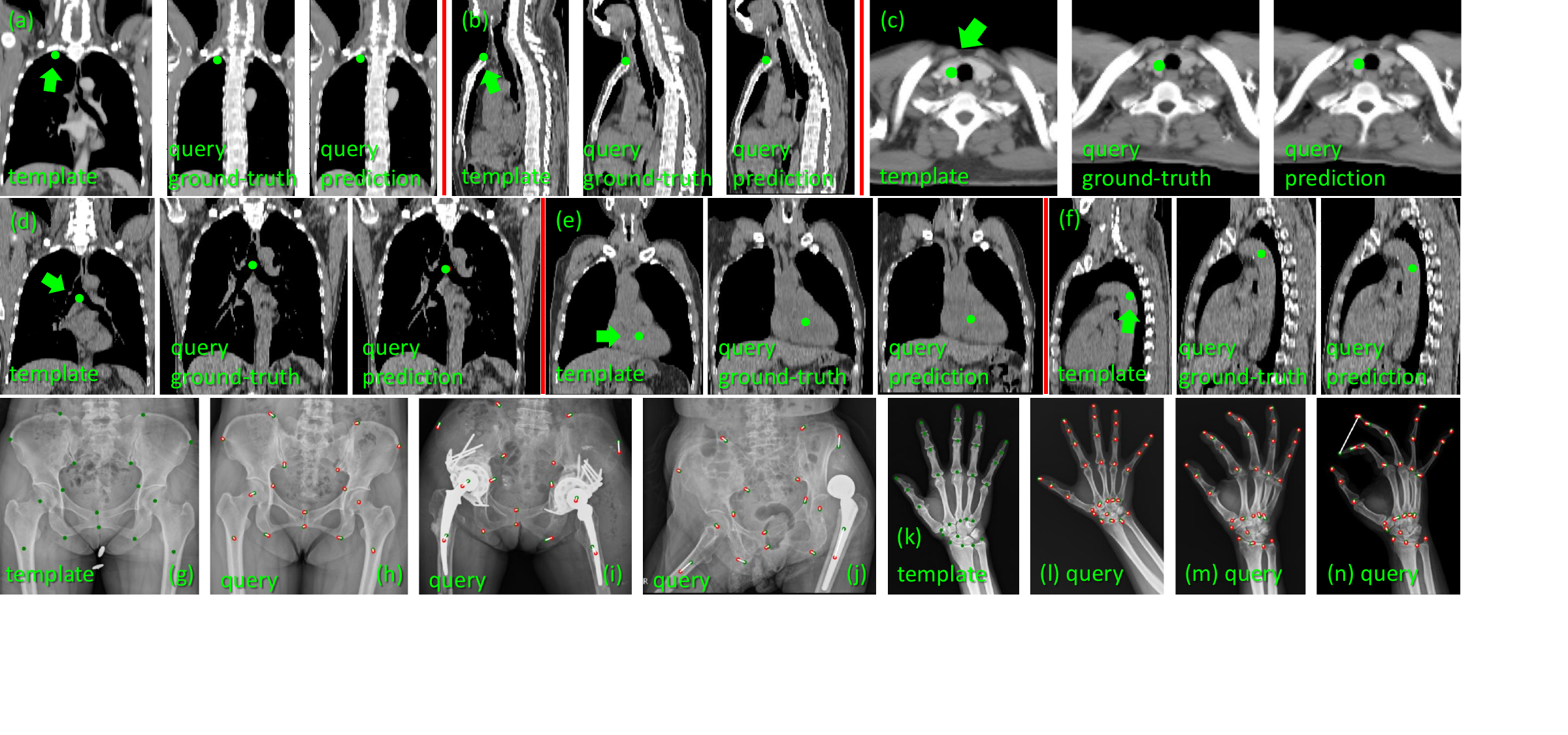} 
	\caption{Examples of landmark detection results of SAM. Each image type (chest CT, pelvic X-ray, and hand X-ray) has only one template image. For CT images (a--f), 
		the views (axial, coronal, sagittal) are selected in order to clearly show the point. 
		For X-ray images (g--n), the green and red points are respectively ground-truth and predicted points, while the white lines show their correspondence. (f), (j), and (n) are failure cases in which the prediction errors are relatively large.} 
	\label{fig:landmark_examples} 
\end{figure*}

\begin{figure*}[]
	\centering
	\includegraphics[width=\textwidth,trim=0 240 100 0, clip]{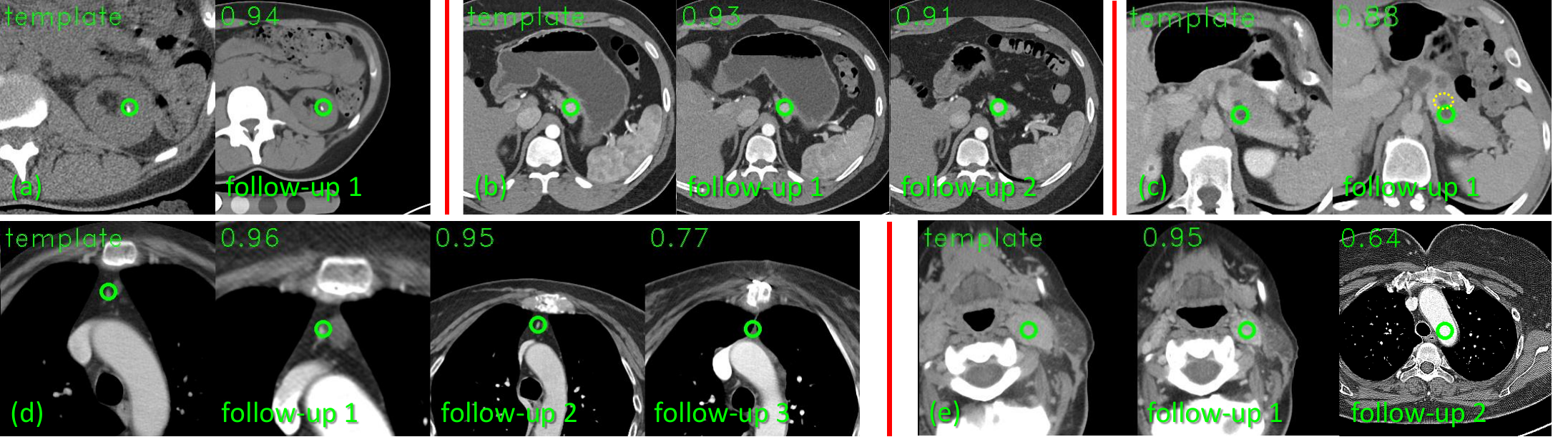} 
	\caption{Examples of lesion matching results of SAM on DeepLesion. 
		Green circles mark the lesion's central point in the template scan and the detected points in follow-up scans. The predicted similarity scores are also shown. See text for details.
	} 
	\label{fig:lesion_match_examples} 
\end{figure*}

\subsection{Tasks and Datasets}
\label{sec:exp:data}

For the CT-based tasks, we train a unified SAM model using the combination of three CT datasets: DeepLesion, NIH-Lymph Node (NIH-LN), and an in-house chest CT dataset (ChestCT). DeepLesion~\cite{Yan2018DeepLesion, Yan2018graph} is a large CT dataset with 20K sub-volumes covering the whole body. It contains a variety of contrast phases, reconstruction protocols, and pathology. NIH-LN~\cite{NIH_LN_dataset} includes 176 chest-abdomen-pelvis CT scans. ChestCT contains 94 patients, each with a contrast-enhanced (CE) and a non-contrast (NC) scan that are pre-aligned. When training SAM, no labels from the three datasets are used. We follow the official data split for DeepLesion, while use all images in NIH-LN for training. For ChestCT, we randomly split the patients to 65 training, 10 validation, and 19 testing. 

{\bf 3D landmark detection} is tested using ChestCT, which contains manual annotations of 35 organs for each patient. We calculated 19 landmarks to evaluate SAM, such as the top of left lung, the bifurcation of trachea, etc. 
To detect landmarks in the one-shot setting, we need to select a template image from the training set. The sample whose landmarks are closest to the average normalized landmarks is chosen as the template~\cite{Wang2020LTNet}.

{\bf 2D landmark detection} is conducted on X-rays following the settings in~\cite{Li2020landmark}. The hand and pelvic models are trained separately. The hand X-ray dataset contains 378 images for training and 93 for testing. The images resolutions are in $\sim 1500\times 2000$. 30 landmarks are manually labeled in each image. The pelvic X-rays are taken on each patient's pelvic bone with resolutions of $\sim 2500\times 2000$. 800 images are used for training and 200 for testing. The template images are chosen in the same way as in ChestCT. In both hand and pelvic X-rays, there are high structural and appearance variations, caused by various poses and pathologies
. Landmarks in X-rays are useful in bone age estimation, arthritis diagnosis, etc.

{\bf 3D Lesion matching} is tested using DeepLesion, which includes 32K lesions annotated on 4K patients. Most patients have multiple follow-up CTs. Given a lesion annotation on one CT, the task of lesion matching is to locate the corresponding lesion on another CT of the same patient. We manually annotate 3,556 lesion pairs on the validation and test sets of DeepLesion. 
We extract SAM embedding from the center of the box of the template lesion, and use it to detect a matched point in the follow-up scan.

{\bf Evaluation metrics:} We use the mean radial error (MRE)~\cite{Li2020landmark} and maximum radial error as the evaluation metrics. Radial errors are the Euclidean distances of predicted and ground-truth coordinates. 
For lesion matching, the central points of the lesions' 3D bounding-boxes are used as landmarks to compute errors. Besides, if the matched point falls in the lesion's bounding-box, it is regarded as a correct match to calculate an accuracy.

\subsection{Implementation Details}
\label{sec:exp:implement}

SAM is implemented in PyTorch with the help of data augmentation libraries~\cite{Fabian2020batch, Perez-Garcia2020torchio}. It is trained using rectified Adam (RAdam)~\cite{liu2019radam} with a learning rate of 0.0001. 
The number of patch pairs per batch is 8 for CT and 16 for X-ray. All models converge in about 25K iterations. We normalize CT images to an isotropic spacing of $ 2\text{mm} $, whereas the X-ray images are resized to $ 512\times512 $ as in \cite{Li2020landmark}. Following \cite{Li2020landmark}, the detected landmarks on resized X-rays are mapped back to match the original image size, and then errors are computed. During training, the cropped patch size is $ 32\times96\times96 $ for CT and $ 400\times400 $ for X-ray. During inference, the entire image without cropping is sent to the network. We have verified that the difference in input size between training and inference does not affect performance. During training, suppose the side length of the training patch is $ b $, the side length of the overlapped region between two patches is $ a $. We make the overlap ratio $ a/b $ to be a random value between 0.2 and 1. To promote reproducibility, we illustrate the detailed 3D and 2D network architectures in \Fig{network}. The backbone network is 3D ResNet-18 for CT and 2D ResNet-34 for X-ray. Further increasing the network depth did not result in significant improvements. 
The global and local embedding dimensions are both $ c=128 $. The default sampling hyper-parameters for CT are $ \npos=100, \nneg=500, \ngrand=1000$, and $ \nlcand=2\times10^4 $. For X-ray, they are the same except for $ \nlcand=5000 $ as we hope the local embedding to be more locally focused in 2D images. We will study the selection of the hyper-parameters and show that SAM is robust to their values in \Sec{exp:ablation}.

\subsection{Qualitative Results}
\label{sec:exp:qualitative}

As shown in \Fig{landmark_examples}, SAM is able to accurately detect various landmarks in different body parts with only one template image. 
A failure case is shown in (f), where the detection of the top of the descending aorta is off in the $ z $-axis
, which is mainly because the tube-shaped aorta does not have discriminative textures along the $ z $-axis. For X-ray images, SAM can locate the landmarks in the presence of body rotation, deformation, and metal prostheses. Failure cases occur when there are severe pathological deformations that significantly differ from the template image (subplots (j) and (n)). 

SAM can match a variety of lesions effectively in follow-up CTs, see \Fig{lesion_match_examples}. In (a), it locates a tiny kidney lesion despite the different noise levels in two images. In (b), SAM detects the lesion in ``follow-up 2'' even though the shape of its surrounding organ has changed. In (c), the algorithm successfully differentiates the true matched lesion from an adjacent one in the dashed yellow circle. (d) is an interesting case where the lesion disappeared in ``follow-up 3'', possibly due to surgery. SAM can locate the correct position even if the lesion no longer exists, showing that it did not simply match the lesion texture, but 
has learned the body part from the anatomical context. Besides, ``follow-up 1'' in (d) is a CT scan with a limited field of view, which SAM handled well. Sometimes a lesion in the template scan may not exist in the follow-up scan due to the scanning range differences in the $z$-axis, which is the case in ``follow-up 2'' of (e). SAM detected an unrelated point 
with a low similarity score, which can be filtered with a score threshold when used in practice.

\subsection{Quantitative Results}
\label{sec:exp:compare}

\begin{table}[]
	\begin{center}
		\caption{Comparison of methods on the ChestCT dataset.}
		\label{tbl:chestCTcmp} 
		\small
		\setlength{\tabcolsep}{2pt}
		\renewcommand{\arraystretch}{1.2}
		\begin{tabular}{lccccc}
			\noalign{\smallskip}\hline	
			& MRE$_{\text{CE}}$	& Max$_{\text{CE}}$	& MRE$_{\text{NC}}$	& Max$_{\text{NC}}$	& Time (s) \\
			\hline
			Affine~\cite{Klein2010elastix}	& 8.4$\pm$5.2	& 32.9	& 8.5$\pm$5.3	& 33.1	&  6.82\\
			FFD~\cite{Rueckert1999FFD}	& 5.4$\pm$4.1	& 29.1	& 5.5$\pm$3.9	& 24.6	&  86.66\\
			SyN~\cite{Avants2008ANTS}	& 6.0$\pm$5.2	& 35.1	& 6.0$\pm$4.7	& 29.3	& 78.58 \\
			DEEDS~\cite{heinrich2013mrf}	& 4.6$\pm$3.3	& 18.8	& 4.7$\pm$3.4	& 24.4	&  50.16\\
			VoxelMorph~\cite{Balakrishnan2019VoxelMorph}	& 7.3$\pm$3.6	& 20.1	& 7.4$\pm$3.7	& 20.2	&  7.43\\
			SAM	& \bf 4.3$\pm$3.0	& \bf 16.4	& \bf 4.5$\pm$3.0	& \bf 18.5	& \bf 0.23 \\
			\hline
			SCN (supervised)~\cite{Payer2019landmark}	& 2.6$\pm$2.2	& 15.9	& 2.7$\pm$2.3	& 16.0	& 4.89 \\
			\noalign{\smallskip}\hline
			\multicolumn{6}{p{248pt}}{CE: contrast-enhanced; NC: non-contrast; Template and query images are in the same modality (CE or NC).}\\
			\multicolumn{6}{p{248pt}}{MRE: mean radial error $ \pm $ std.; Max: maximum radial error. Errors are calculated in pixels. The mean, std., and maximum values are computed over results of all landmarks on all test images.}
			\vspace{-3mm}
		\end{tabular}
	\end{center}
\end{table}

In this section, we compare the performance of SAM with existing registration, landmark detection, and lesion matching algorithms. We also display the results of fully supervised methods trained on all labeled samples as an upper bound for each task. Note that the fully supervised methods should not be directly compared with SAM since they use significantly more training labels on predefined landmarks.

For the 3D landmark detection task, we compare with four widely used conventional registration methods~\cite{Klein2010elastix,Rueckert1999FFD,Avants2008ANTS,heinrich2013mrf} and the unsupervised deep learning-based method VoxelMorph~\cite{Balakrishnan2019VoxelMorph} in one-shot (single atlas) setting. VoxelMorph is initialized using affine registration in Elastix~\cite{Klein2010elastix}. The template CT image in the training set is registered respectively to the images in the test set. The landmark detection errors are shown in \Table{chestCTcmp}. Free-Form Deformation (FFD)~\cite{Rueckert1999FFD} and Symmetric Normalization (SyN)~\cite{Avants2008ANTS} are the top-performing methods in registering chest CTs~\cite{murphy2011evaluation}, while DEEDS~\cite{heinrich2013mrf} performs the best in a recent survey~\cite{xu2016evaluation} and is shown to be more accurate than a recent deep learning-based approach~\cite{Liu2020JSSR}. SAM consistently outperforms all these methods in all metrics 
while taking only 0.23s to process one volume (including embedding extraction and matching on GPU). VoxelMorph performs better than Affine but inferior to some other deformable conventional methods, which is consistent to the findings in~\cite{Heinrich2020highly, Liu2020JSSR}. Deformable registration methods require a robust initial alignment to perform well, and may struggle when the two images exhibit large differences in terms of body size/pose, organ shape and respiratory status~\cite{xu2016evaluation}. In contrast, SAM is able to encode semantic and fine-grained anatomical information, which addresses the landmark matching problem effectively from a novel perspective. We further use a CE template to detect landmarks in NC images, and vice versa. SAM's MREs are 4.5 and 4.8 (DEEDS: 4.7, 4.7). They are comparable to the results using the same modality to train and test (4.5 and 4.3), showing SAM's robustness to contrast changes in CT. We have also run paired $t $-test between SAM and other methods on the CE task. SAM is significantly better than Affine, FFD, SyN, and VoxelMorph ($ p < 10^{-9} $) while comparable with DEEDS ($ p = 0.05 $) despite being much faster. In addition, SAM also has smaller standard deviation and maximum radial error than other methods, indicating that it is more stable and works better in worst cases.

\begin{table}[]
	\begin{center}
		\caption{Results on the X-ray datasets. Supervised methods are trained in few-shot and fully supervised settings (378 and 800 for hand and pelvic datasets, respectively).}
		\label{tbl:Xraycmp}
		\small
		\setlength{\tabcolsep}{4pt}
		\renewcommand{\arraystretch}{1.2}
		\begin{tabular}{lp{1.6cm}cc}
			\noalign{\smallskip}\hline	
			Method	& \# Labeled samples	& Hand MRE	& Pelvic MRE \\
			\hline
			HRNet~\cite{Sun2019HRNet}	& 5	& 43.3$\pm$112.0	& 139.1$\pm$327.5 \\
			& 50	& 14.1$\pm$14.7	& 35.1$\pm$78.1 \\
			& All	& 12.8$\pm$6.1	& 24.8$\pm$20.0 \\
			DAG~\cite{Li2020landmark}	& 5	& 16.4$\pm$39.5	& 35.3$\pm$53.2 \\
			& 50	& 6.2$\pm$4.2	& 21.5$\pm$19.8 \\
			& All	& \bf 5.6$\pm$3.6	& \bf 18.4$\pm$17.7 \\
			SCN~\cite{Payer2019landmark}	& All	& 6.1$\pm$4.0	& 21.0$\pm$21.9 \\
			SAM	& 1	& 13.1$\pm$32.3	& 32.2$\pm$26.2 \\
			\noalign{\smallskip}\hline
		\end{tabular}
	\end{center}
	\vspace{-3mm}
\end{table}

For the 2D X-ray landmark detection tasks, we compare with a strong supervised baseline HRNet~\cite{Sun2019HRNet} and two state-of-the-art methods, Deep Adaptive Graph (DAG)~\cite{Li2020landmark} and SpatialConfiguration-Net (SCN)~\cite{Payer2019landmark}. We test both few-shot and fully supervised settings. In \Table{Xraycmp}, SAM outperforms HRNet trained on 50 samples (HRNet-50), and DAG-5 in average error. One possible concern is the relatively large variance, which is mainly caused by large errors of a few mismatched outliers (\Fig{landmark_examples} (j)(n)) and the large size of X-ray images ($\sim$2000 px). To verify the significance of the results, we conduct paired $ t $-test between SAM and other methods. In hand X-rays, SAM is significantly better than HRNet-5, HRNet-50, and DAG-5 ($ p  < 0.05$). In pelvic X-ray, SAM is significantly better than HRNet-5 ($ p < 10^{-71} $), but not significant compared to HRNet-50 ($ p = 0.30 $) and DAG-5 ($ p = 0.09 $). Nevertheless, SAM only used one labeled template compared to DAG using five. 
Supervised methods are dedicated to modeling specific landmarks, thus may take advantage of the labeled samples better than self-supervised methods. However, they can only detect what has been \emph{a priori} labeled, while SAM can detect arbitrary anatomical locations.

\begin{table}[]
	\begin{center}
		\caption{Results on the lesion matching task.}
		\label{tbl:lesion_match}
		\small
		\setlength{\tabcolsep}{4pt}
		\renewcommand{\arraystretch}{1.2}
			\begin{tabular}{lcccc}
				\noalign{\smallskip}\hline	
				Method	& Lesion emb.~\cite{Yan2018graph}	& LesaNet~\cite{Yan2019Lesa}	& DLT~\cite{Cai2021DLT}	& SAM \\
				\hline
				Accuracy (\%)	& 80.7	& 82.1	& 82.1	& \bf 91.1 \\
				\noalign{\smallskip}\hline
			\end{tabular}
		\end{center}
		\vspace{-3mm}
	\end{table}

For the lesion matching task, we assess two supervised lesion embedding methods trained on weak labels~\cite{Yan2018graph} or labels mined from radiological reports~\cite{Yan2019Lesa}. Since they need the bounding-box of the lesion as input, we use a lesion detector VULD~\cite{Cai2020VULD} to first detect all lesion candidates in the query CT, and then match them with the template lesion. As shown in \Table{lesion_match}, the accuracies of lesion embedding and LesaNet are both significantly lower than SAM. The bottleneck of these detection+matching methods is detection, since it is difficult to find all lesion candidates (especially when they are small and subtle). The detection recall is only 87.3\% in our experiment. Therefore, our direct matching strategy is more suitable by avoiding the intermediate detection step. We also compare against a recent lesion tracking algorithm, i.e., Deep Lesion Tracker (DLT)~\cite{Cai2021DLT}. DLT first initializes the lesion location in the query CT using image registration, then refines the location using a supervised Siamese network. Its accuracy is 9\% lower than SAM on the 3,556 lesion pairs. Failing to accurately locate some small lesions is the main reason for DLT's errors. DLT was trained on lesions from the official annotations of DeepLesion, whose size distribution is larger than our new test set. This further indicates that supervised methods are limited by their training labels. 

\begin{table}[]
	\begin{center}
		\caption{Comparison of methods on the registration task~\cite{Liu2021SAME}. For ChestCT, within-phase (CE-CE) and cross-phase (CE-NC) registration are tested. Dice scores (\%) are reported.}
		\label{tbl:registration_cmp}
		\small
		\setlength{\tabcolsep}{2pt}
		\renewcommand{\arraystretch}{1.2}
		\begin{tabular}{lcccc}
			\noalign{\smallskip}\hline	
			\multirow{2}{*}{Method}  & \multicolumn{3}{c}{\emph{ChestCT dataset}} & \multirow{2}{1.5cm}{\emph{Abdominal CT dataset}} \\
			& \multicolumn{1}{c}{CE-CE} &
			\multicolumn{1}{c}{CE-NC} &  \multicolumn{1}{c}{Time (s)} & \\
			\hline
			Affine~\cite{Klein2010elastix}	& 28.4	& 28.0	& 3.4	& 21.2 \\
			FFD~\cite{Rueckert1999FFD}	& 49.4	& 48.2	& 93.5	&  33.4 \\
			ANTs (SyN)~\cite{Avants2008ANTS}	& 49.8	& 48.0	& 74.3	& 28.4 \\
			DEEDS~\cite{heinrich2013mrf}	& 52.7	& \bf 51.2	& 45.4	& 46.5 \\
			VoxelMorph~\cite{Balakrishnan2019VoxelMorph}	& 42.7	& 40.8	& 4.0	& 28.5 \\
			\hline
			SAM-affine (proposed)~\cite{Liu2021SAME}	& 33.8	& 33.8	& 0.5	& 29.1 \\
			SAME (proposed)~\cite{Liu2021SAME}	& \bf 54.4	& 51.0	& 1.2	& \bf 49.8 \\
			\hline
			{Supervised~\cite{Cicek20163DUnet,Hering2021reg}}	& 79.7	& 74.1	& 109.3	& {69} \\
			\noalign{\smallskip}\hline
		\end{tabular}
	\end{center}
\end{table}

For the registration task, we conduct atlas-based segmentation on two datasets: ChestCT (35 annotated organs on 90 test image pairs) and an abdominal CT dataset (13 organs on 45 test pairs)~\cite{xu2016evaluation}. Average Dice scores and time cost are reported in \Table{registration_cmp}. First, SAM-based affine registration significantly surpasses traditional optimization-based affine registration in accuracy and speed, suggesting it is a promising registration initialization method. Our proposed SAME (SAM-affine + SAM-coarse + SAM-VoxelMorph)~\cite{Liu2021SAME} outperforms other methods on most metrics, while only taking 1.2s. It benefits from SAM's fast and accurate point matching results and the semantic information contained in the SAM embeddings. 
See more details in~\cite{Liu2021SAME}. 
{For the supervised upper bound, we used the 3D U-Net~\cite{Cicek20163DUnet} for ChestCT and the best supervised registration result of the Learn2Reg challenge~\cite{Hering2021reg} for the abdominal dataset.} Note that all other registration algorithms are unsupervised. 3D U-Net splits the whole volume to overlapped patches to inference separately for better accuracy, which induces more time overhead.

Besides these applications, SAM can potentially be used to aid other medical image analysis tasks. For example, we have proposed a framework to classify CT contrast phases in~\cite{Yan2021Contrast}. We first apply SAM to detect contrast-related landmarks such as the aorta using template point annotations from only one patient. Then we extract Hounsfield Unit values from the landmarks as features and finally train an SVM for classification. This simple method outperforms a deep learning baseline~\cite{Zhou2019Contrast} (F1 95.9\% vs.~94.5\%) while being much more interpretable.

\subsection{Ablation and Parameter Study}
\label{sec:exp:ablation}

\begin{table*}[th!]
	\begin{center}
		\caption{Ablation study of the proposed method. The best result in each metric is shown in bold. CE: contrast-enhanced; NC: non-contrast; MRE: mean radial error$ \pm $std.; Max: maximum radial error. Errors are calculated in pixels.}
		\label{tbl:ablation}
		\small
		\setlength{\tabcolsep}{3pt}
		\renewcommand{\arraystretch}{1.2}
		\begin{tabular}{lccccccccc}
			\noalign{\smallskip}\hline\noalign{\smallskip}
			\multirow{2}{*}{Method}	& \multicolumn{2}{c}{\textit{ChestCT CE}}	& \multicolumn{2}{c}{\textit{ChestCT NC}}	& \textit{Hand X-ray}	& \textit{Pelvic X-ray}	& \multicolumn{3}{c}{\textit{Universal lesion matching}}  \\
			& MRE	& Max	& MRE	& Max	& MRE	& MRE	& Accuracy (\%)	& MRE	& Max \\
			\noalign{\smallskip}\hline\noalign{\smallskip}
			
			(a) SAM (proposed)	& \bf 4.3$\pm$3.0	& \bf 16.4	& \bf 4.5$\pm$3.0	& 18.5	& \bf 13.1$\pm$32.3	& \bf 32.2$\pm$26.2	& 91.1	& \bf 2.7$\pm$2.5	& 28.8 \\
			(b) w/o coarse-to-fine structure	& 4.9$\pm$3.0	& 17.4	& 5.2$\pm$4.3	& 61.2	& 26.1$\pm$70.0	& 34.4$\pm$31.6	& 84.1	& 3.5$\pm$3.1	& 41.1 \\
			(c) Test: Global embedding only	& 9.9$\pm$12.1	& 80.1	& 9.4$\pm$10.4	& 79.3	& 437.0$\pm$431.1	& 48.8$\pm$28.8	& 52.5	& 6.6$\pm$3.2	& 29.9 \\
			(d) Test: Local embedding only	& 6.3$\pm$12.2	& 150.1	& 6.5$\pm$9.2	& 84.7	& 473.4$\pm$519.2	& 34.6$\pm$43.3	& 90.1	& 3.2$\pm$6.6	& 130.4 \\
			
			\hline
			(e) w/o hard negative sampling	& 4.5$\pm$3.2	& 17.6	& 4.7$\pm$3.2	& \bf 16.6	& 19.1$\pm$46.3	& 39.7$\pm$33.0	& 89.0	& 2.9$\pm$2.7	& 26.9 \\
			(f) w/o diverse negative sampling	& 5.3$\pm$6.6	& 58.7	& 7.6$\pm$11.6	& 83.4	& 14.7$\pm$39.5	& 45.1$\pm$41.1	& \bf 92.2	& 2.7$\pm$3.4	& 58.3 \\
			\hline					
			(g) w/o cutting FPN connection	& 4.7$\pm$5.3	& 60.2	& 5.9$\pm$9.1	& 79.9	& 13.9$\pm$22.2	& 37.2$\pm$28.2	& 88.6	& 3.0$\pm$2.9	& 44.8	\\
			(h) w/o ImageNet pretraining	& 4.6$\pm$3.2 	& 18.7	& 4.8$\pm$3.2 	& 18.7	& 17.6$\pm$37.6	& 38.1$\pm$29.0	& 88.0	& 3.0$\pm$2.6 	& \bf 26.8 \\
			(i) U-Net backbone~\cite{Cicek20163DUnet, Navab2015UNet}	& 5.9$\pm$5.8 	& 49.3	& 6.0$\pm$5.6 & 	47.9	& 37.2$\pm$74.2	& 41.2$\pm$42.6	& 84.7	& 3.5$\pm$3.4 	& 42.2 \\
			(j) SAM trained on NIH-LN alone	& 4.5$\pm$3.3	& 20.8	& 4.9$\pm$3.7	& 24.2 & -- & -- & 91.4	& 2.7$\pm$2.6 & 30.7 \\
			\noalign{\smallskip}\hline	
		\end{tabular}
	\end{center}
\end{table*}

\begin{figure}[]
	\centering
	\includegraphics[width=\columnwidth,trim=0 350 300 0, clip]{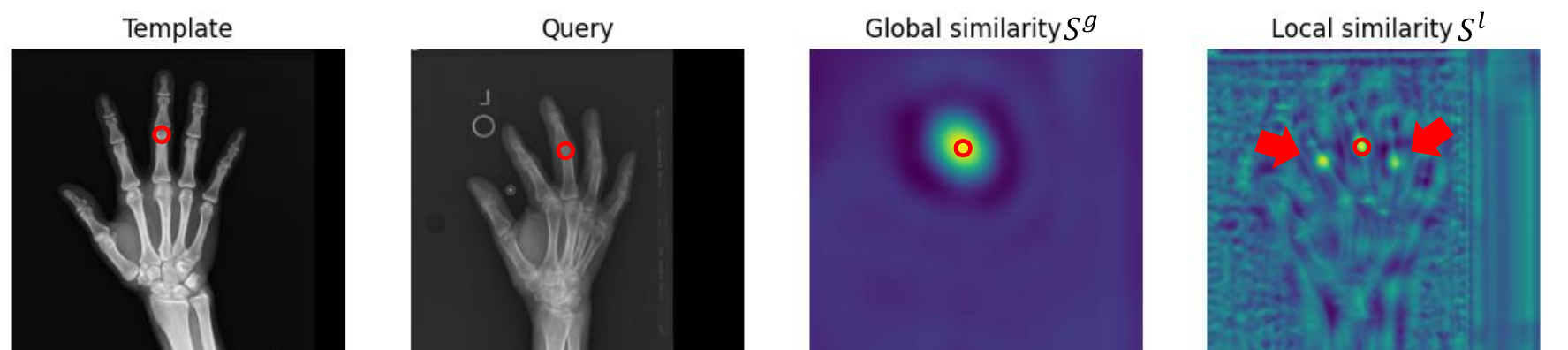} 
	\caption{An example in the hand X-ray dataset. Red circles show the ground-truth landmarks. Arrows indicate spurious highlights in the local similarity map of the query image.} 
	\label{fig:hand_sim}
\end{figure}

{\bf Coarse-to-fine embedding architecture:}
\Table{ablation} outlines the ablation study results on the five tasks (landmark detection on ChestCT CE, ChestCT NC, hand X-ray, pelvic X-ray; and universal lesion matching). As we can see, the coarse-to-fine embedding learning strategy is important to SAM. For instance, we test only learning one fine-scale embedding for each pixel and remove the global embedding, which is similar to other dense self-supervised methods~\cite{Pinheiro2020dense, Wang2021Dense}. As shown in row (b), the results degrade in all tasks. In this setting, the fine-scale embedding has to encode both global and local anatomical information, which is more challenging than our coarse-to-fine strategy. We also train both global and local embeddings, but only use one of them for inference. As shown in rows (c) and (d), the results are not promising. The global embedding can provide a rough localization, but it is not locally accurate (see the small offsets between the global similarity peaks and the red points in Fig.~\ref{fig:inference}). When using it alone, mean accuracies of all tasks drop considerably. The local embedding can localize more precisely, but it may also highlight distant areas with similar local textures if the global similarity is not considered, leading to a large maximum error, which will decrease the algorithm's stability when it is used in practice. This can be further illustrated in \Fig{hand_sim}. Neighboring finger joints display similar local texture and may be mismatched by the local embedding, explaining the large error of this dataset in row (d).

{\bf Sampling strategies:} Removing the hard negative sampling strategy lowers average accuracy in all tasks, as shown in row (e). The difference is more evident in the two X-ray datasets, where the mean errors increase from 13.1 and 32.2 to 19.1 and 39.7, respectively. This is possibly because the appearance of each pixel is less distinctive in X-ray images compared to CT images (\Fig{landmark_examples}), so hard negative pixel sampling is more important for X-rays. 
Row (f) outlines performance when removing the diverse negative sampling strategy and only sampling hard negative pixels in training. We can find that the mean errors increase in the landmark detection tasks, and the maximum errors increase considerably. This is possibly because the model overfits to the hard examples in the training set, so it produces more distant false highlights in the test set, reducing the generalizability and stability of the algorithm.

\begin{table}[]
	\begin{center}
		\caption{MRE with different data augmentation used in training. We start with random crop and add more augmentations step-by-step. Def.~\&~Rot.: deformation and rotation.}
		\label{tbl:augment}
		\small
		\setlength{\tabcolsep}{3.3pt}
		\renewcommand{\arraystretch}{1.2}
		\begin{tabular}{p{2.1cm}p{1cm}p{1cm}p{1cm}p{1cm}p{1cm}}
			\noalign{\smallskip}\hline\noalign{\smallskip}
			Method	& ChestCT CE	& ChestCT NC	& Hand X-ray	& Pelvic X-ray	& Lesion match. \\
			\hline
			Crop only	& 4.7	& 4.7	& 26.7	& 32.6	& 2.8 \\
			+ Scaling	& \bf 4.3	& 4.6	& 21.9	& 32.9	& 2.7 \\
			+ Intensity jitter	& \bf 4.3	&\bf  4.5	& 22.6	& 33.8	& 2.7 \\
			+ Def.~\&~Rot.	& 4.5	& 4.6	& \bf 13.1	& \bf 32.2	& \bf 2.6 \\
			+ Flip	& 4.6	& 5.0	& 64.9	& 326.4	& 2.8 \\
			\noalign{\smallskip}\hline	
		\end{tabular}
	\end{center}
\end{table}

{\bf Backbone and training strategies:} The backbone of SAM is a customized FPN initialized with ImageNet pretrained weights using the inflated 3D technique~\cite{Carreira2017I3D}. The connection between the coarsest FPN feature and its upper level is cut to make global and local embeddings more independent, forcing the local embeddings to focus on fine-level features. In row (g) of \Table{ablation}, we do not cut the connection and observe accuracy drop in all datasets. In row (h), the backbone is randomly initialized and trained for twice as many epochs. The accuracy dropped slightly, indicating that ImageNet initialization is helpful. We have also tried other backbone choices such as 3D and 2D U-Nets~\cite{Cicek20163DUnet, Navab2015UNet} with a maximum channel number of 512 and a longer training schedule than SAM to compensate for random initialization. However, the accuracy of U-Net is not as good as SAM with random initialization (row (i) vs.~(h)), demonstrating the superiority of the proposed backbone. In row (g), we train SAM on the NIH-LN dataset alone and test it on DeepLesion and ChestCT. Its performance only degrades slightly compared to row (a), which uses DeepLesion and ChestCT in training. This result indicates that the embeddings learned in SAM are generalizable across datasets. It is also thanks to the fact that the intrinsic structure of human organs are relatively stable in radiological images, which is the basis of this work.

{\bf Data augmentation:} \Table{augment} analyzes the effect of different data augmentation approaches, which is key to contrastive learning~\cite{Chen2020SimCLR}. Results show that random cropping and scaling suffice to learn good embeddings in CT, while random deformation and rotation (within 30$ ^{\circ} $) is crucial for hand X-ray, as rotated and deformed hands are common in the dataset. Random flipping harms performance as the medical images we used are normalized in anatomical direction. For pelvic X-rays, it is hard to distinguish their left and right sides, so flipped images confuse model training. Color distortion is found helpful for natural images in~\cite{Chen2020SimCLR}. However, intensity jitter brought little gain in our experiments, probably because the intensity distribution is relatively stable in CT and X-ray. We hypothesize that a data augmentation approach for learning effective anatomical embedding is essential only if the corresponding image variance commonly exists in the data. 

\begin{figure*}[t]
	\centering
	\includegraphics[width=.8\textwidth,trim=0 340 240 0, clip]{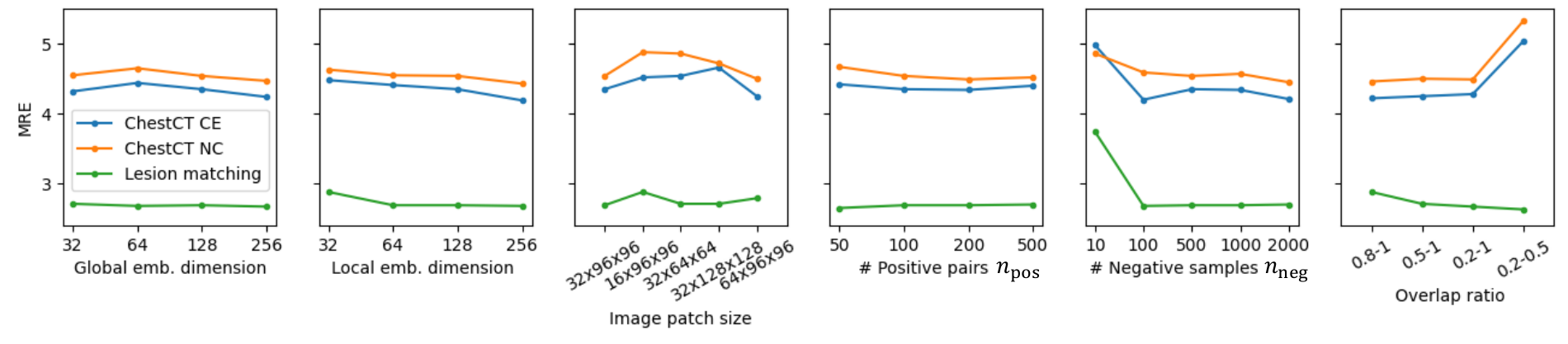} 
	\caption{Parameter study on the ChestCT and lesion matching datasets. Lower MRE is better.
	} 
	\label{fig:param_study} 
\end{figure*}

{\bf Parameter study:} The influence of parameter values is investigated in \Fig{param_study}. We can see that SAM is robust to embedding dimensions, although larger dimension slightly improves performance~\cite{Chen2020SimCLR}. The size of the image patch in $ z $-axis should not be too small ($ \geq $ 32 pixels). SAM is robust to $ \npos$ and $\nneg $ 
as long as $ \nneg\geq 100 $. 
SAM is robust to the overlap ratio of the two augmented patches in training, but the overlap upper bound should not be too small.

\subsection{Comparison with Other Feature Extractors}
\label{sec:exp:ft_baseline}

In this section, we compare SAM with several feature baselines including a traditional handcrafted descriptor, supervised or self-supervised networks pretrained from other tasks, and random initialization.
BRIEF~\cite{Calonder2010BRIEF} is a handcrafted feature descriptor computed using random intensity difference tests. Med3D~\cite{Chen2019Med3D} is a pretrained 3D encoder network for medical images. We adopt a ResNet-18 model pretrained on 23 datasets\footnote{https://github.com/Tencent/MedicalNet}. Models Genesis~\cite{Zhou2020Genesis} is a self-supervised framework that learns features from unlabeled medical images by image restoration. A released 3D U-Net model trained on chest CT data\footnote{https://github.com/MrGiovanni/ModelsGenesis/tree/master/pytorch} is utilized. {The pretrained models of Med3D and Models Genesis are used to extract feature maps for template and query images and find the matched points as illustrated in \Fig{inference}.} All the features above are matched using L2 distance. Finally, a SAM model with random weights is also compared as a baseline. The results are exhibited in \Table{feature_cmp}. We can see that SAM obtains the best accuracy in all tasks. BRIEF is not suitable for anatomical point matching in radiological images, possibly because it cannot model global information. Furthermore, the local textures in radiological images are not as discriminative as those in natural images, which brings challenges to simple gradient-based feature descriptors. Both Med3D and Models Genesis are designed for network pretraining on medical images. They may have learned useful radiological features for downstream tasks, but they are not optimal for the task of anatomical point matching, so there is a clear performance gap between them and SAM. {Random network (d) produces much higher error in X-ray compared to CT possibly because the image features of anatomical points in X-ray are less distinctive than in CT.}

\begin{table*}[th!]
	\begin{center}
		\caption{Comparison of different feature extraction algorithms.  ``-'' means the model is 3D but the task is 2D.}
		\label{tbl:feature_cmp}
		\small
		\setlength{\tabcolsep}{3pt}
		\renewcommand{\arraystretch}{1.2}
		\begin{tabular}{lccccccccc}
			\noalign{\smallskip}\hline\noalign{\smallskip}
			\multirow{2}{*}{Method}	& \multicolumn{2}{c}{\textit{ChestCT CE}}	& \multicolumn{2}{c}{\textit{ChestCT NC}}	& \textit{Hand X-ray}	& \textit{Pelvic X-ray}	& \multicolumn{3}{c}{\textit{Universal lesion matching}}  \\
			& MRE	& Max	& MRE	& Max	& MRE	& MRE	& Accuracy (\%)	& MRE	& Max \\
			\noalign{\smallskip}\hline\noalign{\smallskip}
			
			(a) BRIEF~\cite{Calonder2010BRIEF}	& 23.3$\pm$35.2 	& 153.2	& 19.1$\pm$28.5 	& 148.1	& 324.9$\pm$376.3 	& 777.0$\pm$618.0	& 62.4	& 20.6$\pm$33.8	& 177.0 \\
			(b) Med3D~\cite{Chen2019Med3D}	& 14.0$\pm$13.8 	& 67.0	& 14.4$\pm$14.1 	& 63.1	& -	& -	& 54.1	& 16.6$\pm$25.1 	& 188.5 \\
			(c) Models genesis~\cite{Zhou2020Genesis}	& 8.7$\pm$8.1 	& 53.3	& 7.7$\pm$7.4 	& 64.7	& -	& -	& 47.5	& 5.6$\pm$7.1 	& 85.2 \\
			(d) SAM with random weights	& 8.7$\pm$10.0 	& 91.3	& 7.8$\pm$8.1 	& 78.2	& 112.0$\pm$158.1	& 320.0$\pm$357.2	& 67.1	& 7.1$\pm$10.2 	& 101.6 \\
			{(e) SAM (proposed)} & {\bf 4.3$\pm$3.0} & {\bf 16.4} & {\bf 4.5$\pm$3.0} & {\bf 18.5} & {\bf 13.1$\pm$32.3} & {\bf 32.2$\pm$26.2} & {\bf 91.1} & {\bf 2.7$\pm$2.5} & {\bf 28.8} \\
			\noalign{\smallskip}\hline	
		\end{tabular}
	\end{center}
\end{table*}

\subsection{Network Initialization from SAM}
\label{sec:exp:net_init}

Network initialization is the purpose of many self-supervised learning works in medical imaging~\cite{Zhou2020Genesis, Zhuang2019rubik, Azizi2021big}. Although our goal is learning anatomical embeddings, we wonder whether SAM can also be used for network initialization. 
We run liver and liver tumor segmentation experiments on the LiTS dataset~\cite{Bilic2019LiTS} to verify this idea. Among the 131 labeled CTs, we randomly select 105 for training and the rest for testing. Another experiment is done by training on only 10 labeled samples to simulate a small training set. The 3D U-Net~\cite{Cicek20163DUnet} with combined cross entropy and Dice loss is used as the segmentation framework. We compare two ways to initialize the U-Net: random, or from a pretrained SAM model. In both ways, we train using RAdam~\cite{liu2019radam} for 500 epochs in the experiment with 105 training samples to make the network converge sufficiently, meanwhile in the experiment with 10 samples we only train for 100 epochs to prevent overfitting. The SAM model is trained using the same method as in \Sec{exp:implement}, except that we change the backbone to 3D U-Net.

From \Table{SAM_pretrain} we can see that initialization from SAM consistently outperforms random initialization, especially when there are fewer training samples. Besides, finetuning from SAM benefits liver tumor segmentation more than liver segmentation, which is possibly because liver tumor segmentation is a harder task and requires more fine-scale local features, while the SAM model already learned good fine-scale features from the pixel discrimination task. These results demonstrate that SAM has the potential to be used for network initialization, although it is not designed for the purpose. It will be an interesting future work to compare SAM with other self-supervised network initialization algorithms.

\begin{table}[]
	\begin{center}
		\caption{Comparison of different network initialization methods. The liver and liver tumor segmentation accuracy on the LiTS dataset~\cite{Bilic2019LiTS} is shown. ASD: average surface distance.}
		\label{tbl:SAM_pretrain}
		\small
		\setlength{\tabcolsep}{3pt}
		\renewcommand{\arraystretch}{1.2}
		\begin{tabular}{c|c|cc|cc}
			\hline
			\multirow{2}{*}{\# Labels}	& \multirow{2}{*}{Initialization}	& \multicolumn{2}{c|}{Liver} & \multicolumn{2}{c}{Live tumor} \\
			\cline{3-6}
			& 						& Dice				& ASD			& Dice				& ASD \\
			\hline
			\multirow{2}{*}{105}	& Random	& \bf 0.97$\pm$0.02	& 1.4$\pm$1.3	& 0.62$\pm$0.13	& 1.9$\pm$2.3 \\
			\cline{2-6}
			& From SAM	& \bf 0.97$\pm$0.03	& \bf 1.3$\pm$1.4	& \bf 0.64$\pm$0.13	& \bf 1.8$\pm$3.0 \\
			\hline
			\multirow{2}{*}{10}		& Random	& 0.86$\pm$0.08	& 3.2$\pm$2.6	& 0.42$\pm$0.34	& 5.7$\pm$4.5 \\
			\cline{2-6}
			& From SAM	& \bf 0.90$\pm$0.04	& \bf 2.1$\pm$3.8	& \bf 0.55$\pm$0.23	& \bf 4.6$\pm$6.9 \\
			\hline	
		\end{tabular}
	\end{center}
\end{table}

\subsection{Incorporating Auxiliary Label Information}
\label{sec:exp:semisup}

Our main goal is to learn universal anatomical embeddings from unlabeled radiological images. Meanwhile, it is possible that a few labeled samples can further improve the landmark detection accuracy of SAM. Thus, we make attempts to extend SAM to the semi-supervised setting. Specifically, when a training sample is unlabeled, the original unsupervised selection strategy will be used to select positive pairs from two augmented image patches. When a training sample contains labeled landmarks, we select the corresponding landmarks from another random labeled image to form a positive pair. In this way, the label information is incorporated, which may help SAM to learn  ``cross-image'' similarity better. We conduct this supervised positive selection strategy for the labeled samples with a probability (0.5 in our experiments), because we find that using supervised selection all the time will cause overfitting and degrade the performance. Results are shown in \Table{semisup}. We observe that SAM does benefit from few-shot labeled samples in both 3D CT and 2D X-ray tasks. When 50 labeled samples are used, SAM is comparable with HRNet trained with all labeled samples (\Table{Xraycmp}). How to leverage label information more effectively will be a future work. Instead of improving SAM with label information, another future direction may be improving landmark detection algorithms with SAM-style self-supervision, such as network initialization (\Sec{exp:net_init}) or a regularization loss term.

\begin{table}[h!]
	\begin{center}
		\caption{MRE of SAM with label information. $ x +$1 means $ x $ labeled samples used in training and 1 used as inference template. 20/50 means 20 for ChestCT and 50 for X-ray.}
		\label{tbl:semisup}
		\small
		\setlength{\tabcolsep}{3pt}
		\renewcommand{\arraystretch}{1.2}
			\begin{tabular}{p{1.6cm}p{1.4cm}p{1.4cm}p{1.6cm}p{1.6cm}}
				\noalign{\smallskip}\hline	
				\# Labeled samples	& ChestCT CE	& ChestCT NC	& Hand X-ray	& Pelvic X-ray \\
				\hline
				0+1	& 4.3$\pm$3.0	& 4.5$\pm$3.0	& 13.1$\pm$32.3	& 32.2$\pm$26.2	\\
				5+1	& 4.1$\pm$2.9 	& 4.2$\pm$2.9 	& 11.7$\pm$25.7	& 31.8$\pm$24.4 \\
				20/50+1	& \bf 3.9$\pm$2.7 	& \bf 4.0$\pm$2.8 	& \bf 9.5$\pm$12.8	& \bf 27.9$\pm$22.6 \\
				\noalign{\smallskip}\hline
			\end{tabular}
		\end{center}
		\vspace{-3mm}
	\end{table}

\section{Conclusion}

In this paper, we propose self-supervised anatomical embedding (SAM) to learn pixel-wise anatomical representation from unlabeled radiological images. It is fast and easy to implement. 
SAM can be used to locate arbitrary body parts with only one labeled template.
The learned embeddings are robust to common image variations, applicable to various image modalities and body parts, and generalizable. 
It can be used in universal landmark detection, lesion matching, and registration. Comprehensive experimental results showed that SAM consistently outperforms other registration algorithms and supervised learning methods in few-shot scenarios. 

\section{Appendix}

\subsection{Universal Anatomical Point Matching in CT}

\begin{figure*}[]
	\centerline{\includegraphics[width=.9\textwidth,trim=0 45 70 0, clip]{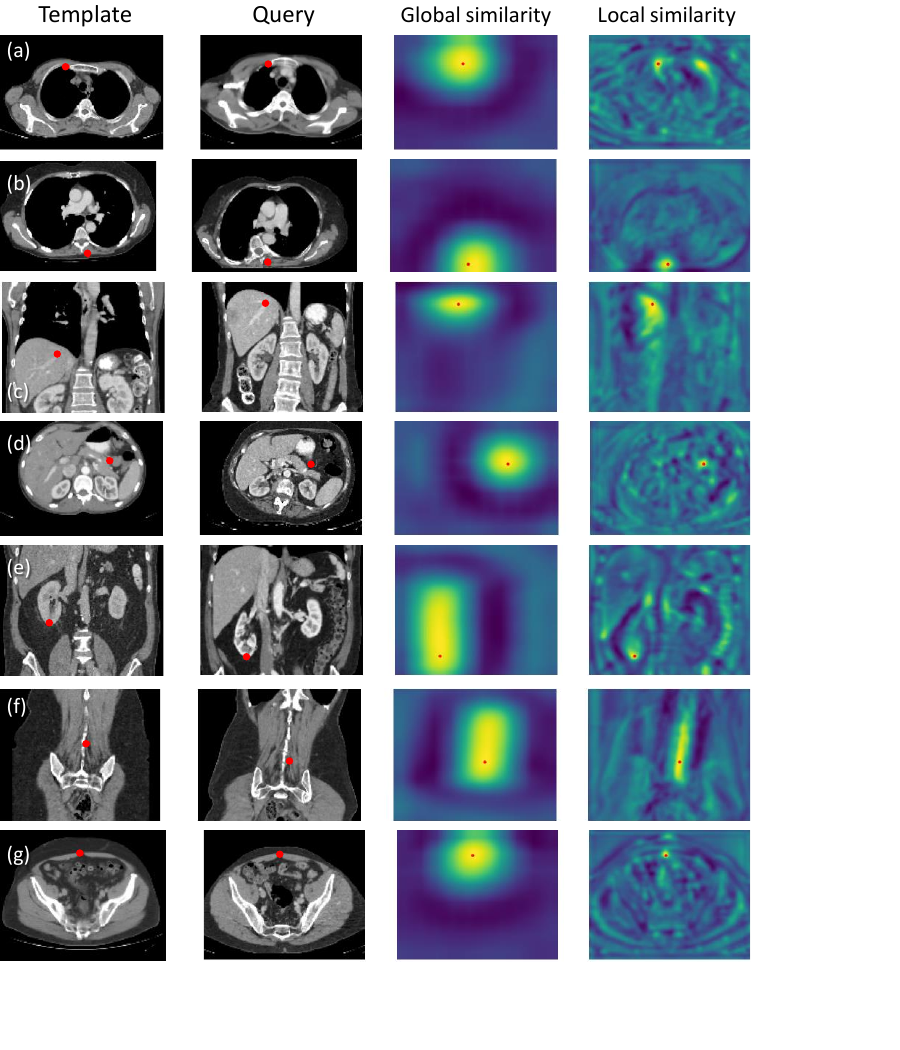}} 
	\caption{Random anatomical point matching results. We randomly select an anatomical point in a template CT image, and then use SAM to find its matched point in a query image from another patient. The template image, query image, global similarity map, and local similarity map are displayed in each row. The red points are selected points in template images or detected points in query images and similarity maps. The views (axial, coronal, sagittal) in each row were selected in order to clearly show the point. (a) Right edge of sternum, (b) left spinal extensors, (c) right hepatic vein, (d) cauda pancreatis, (e) bottom of right kidney, (f) left multifidus muscle, and (g) center of abdomen.}
	\label{fig:rand_match} 
\end{figure*}

To show that SAM can be used to detect arbitrary anatomical locations, we randomly select a point in a template CT image, and then use SAM to find its matched point in a query image from another patient. Examples are demonstrated in \Fig{rand_match}. SAM can accurately find the matched anatomical location in the query image despite significant inter-subject variability, organ deformation, and contrast changes.

\subsection{Template Selection for Landmark Detection}

To detect landmarks in the one-shot setting, we need to select a template image from the training set. We chose the image whose landmarks are closest to the average normalized landmarks as the template, which is similar to the atlas selection method in one-shot segmentation~\cite{Wang2020LTNet}. Suppose the 3D landmark annotations for image $ i $ is $ L_i\in\mathbb{R}^{n_l\times3} $, where $ n_l=19 $ is the number of landmarks. We first normalized each column of $ L_i $ to $ 0\sim 1 $ to get $ \tilde{L}_i $, then computed the average normalized coordinates across samples, $ \bar{L}=\frac{1}{n_\text{train}}\sum_{i=1}^{n_\text{train}}\tilde{L}_i $, where $  n_\text{train} $ is the number of training images. Last, the template image $ \hat{i} $ was selected as $ \hat{i}=\textrm{argmin}_{1\leq i\leq n_\text{train}} \|\tilde{L}_i-\bar{L}\|_F^2 $, where $ \|\cdot\|_F $ is the Frobenius norm. The template images of the 2D X-ray datasets were selected using the same approach.

\subsection{Landmark Computation and Detection on ChestCT}

The ChestCT dataset contains manually annotated masks of 35 organs for each patient. We used the organ masks to compute 19 landmarks. Take ``top of aortic arch'' as an example, we first found the most superior axial slice that contains the mask of ``aortic arch'', and then calculated its center of mass as the landmark. These 19 landmarks include trachea bifurcation, bottom of right/left internal jugular vein, bottom of ascending aorta, top of descending aorta, top/anterior of aorta arch, 3D center of aorta arch, 3D center of heart, left end of left bronchus, left/top of pulmonary artery, 3D center of pulmonary artery, top of left/right lung, 3D center of left/right thyroid, top of sternum, and top of superior vena cava vein.

\begin{figure*}[]
	\centerline{\includegraphics[width=\textwidth]{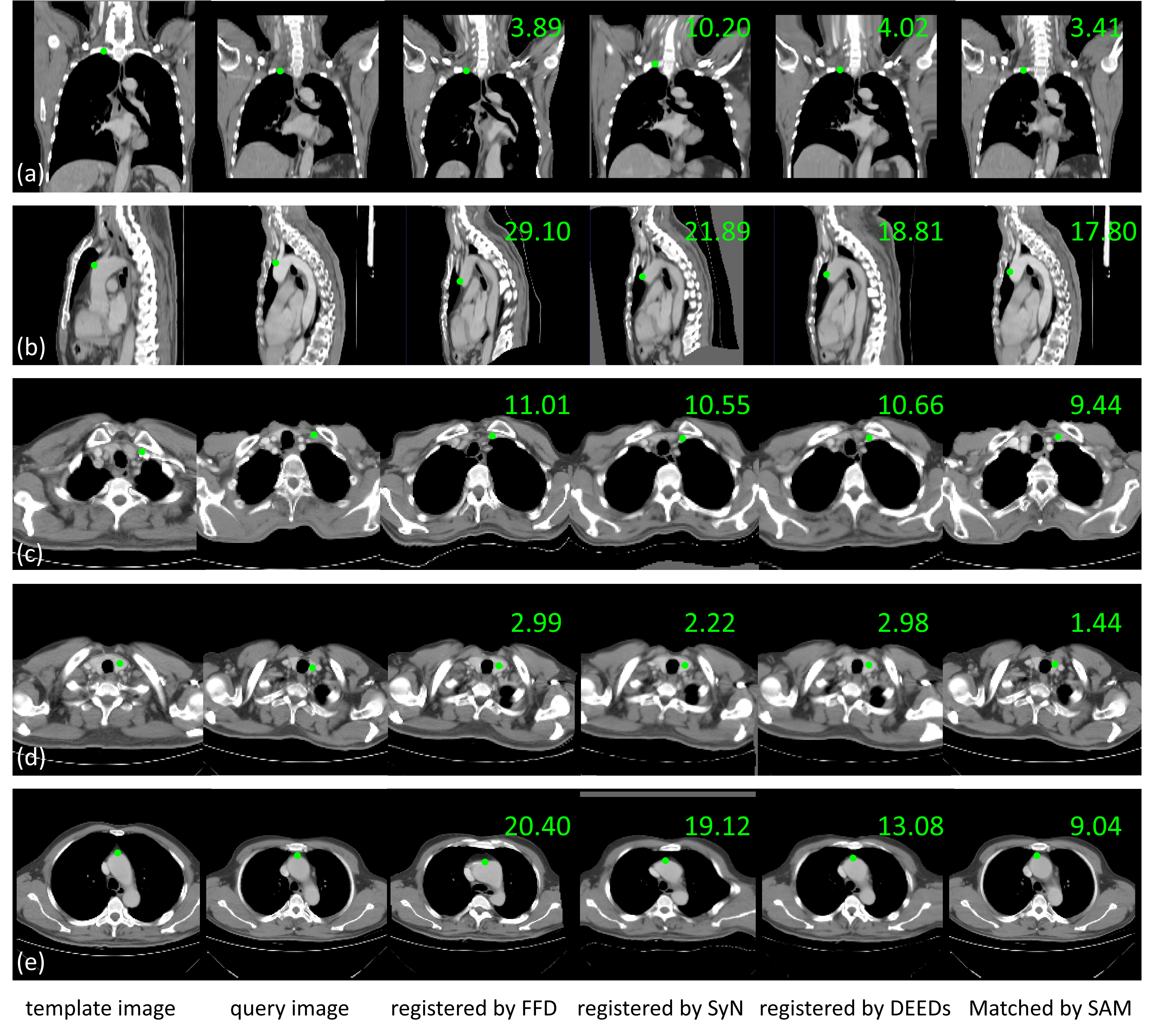}} 
	\caption{3D landmark detection results in chestCT using different methods. The mean radial error \cite{Li2020landmark} of each method is also shown. Columns 3 to 5 present registered/deformed template images according to the query one using FFD \cite{Rueckert1999FFD}, SyN \cite{Avants2008ANTS}, and DEEDs \cite{heinrich2013mrf}, respectively. {\bf Larger appearance inconsistency between the registered image and the query one typically indicates larger landmark detection errors.} {\bf(a)} shows an example for detecting the top of right lung, where lungs in the template image are larger than those in the query one. Although most methods performed well to detect the lung top, registration based methods present obvious deformation errors near the lung bottom, especially for FFD and SyN. {\bf(b)} and {\bf(c)} show examples of detecting the anterior of aorta arch and bottom of internal jugular vein. The query images are quite different from the template in body pose (b) and shape (c), and registration methods exhibit large deformation error near aorta arc (b) and anterior of chest wall (c). {\bf(d)} presents a case that the template and query images have different thyroid sizes, where SAM is able to provide more accurate localization. {\bf(e)} shows an example 
		where the registration methods could not perform well due to the large difference between the template image and the query one, while SAM directly matched the landmark with higher accuracy.}
	\label{fig:ct_landmark}  
\end{figure*}

The 3D landmark detection comparison between SAM and four widely-used conventional registration methods are illustrated in \Fig{ct_landmark}. It can be observed that conventional registration may struggle when the two images exhibit large differences in terms of body size/pose, organ shape and respiratory status, while SAM detected the landmarks more accurately with self-supervised appearance embeddings.

\subsection{Universal Lesion Matching in DeepLesion}

\begin{figure*}[]
	\centerline{\includegraphics[width=\textwidth,trim=0 100 50 0, clip]{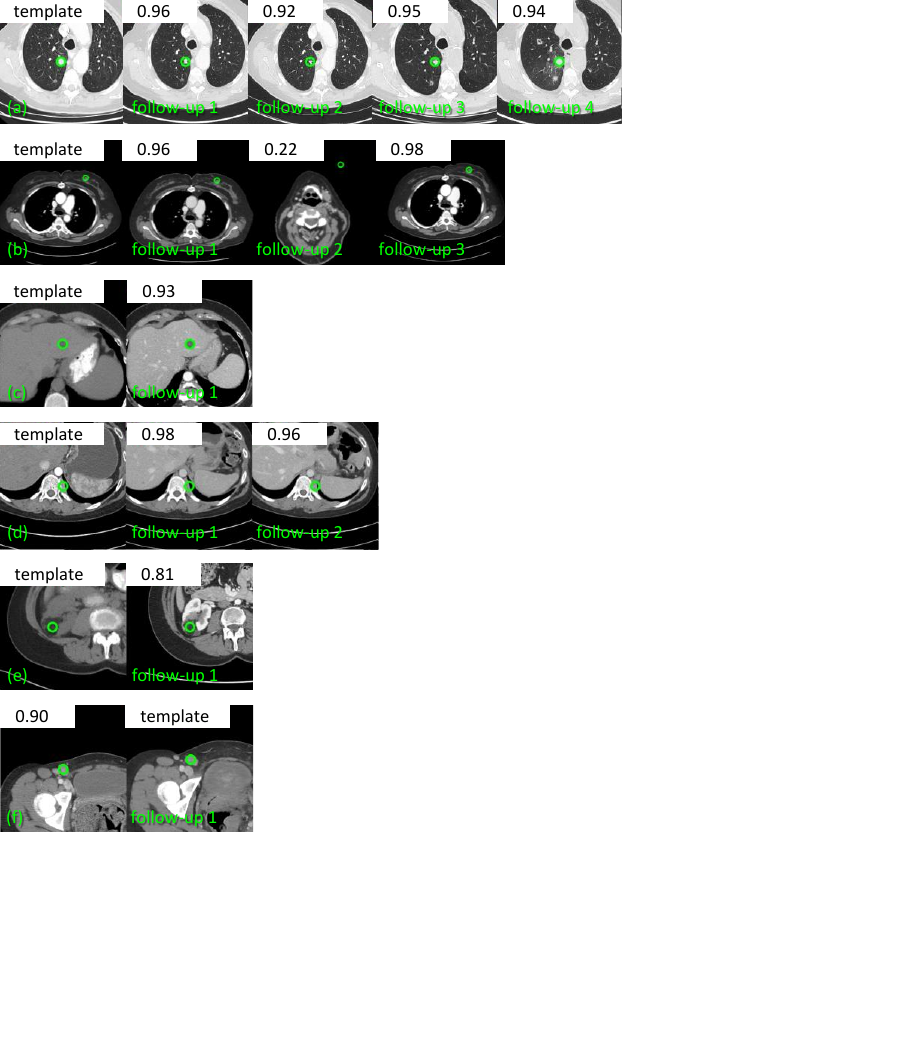}} 
	\caption{Examples of lesion matching results of SAM on DeepLesion.
		Green circles mark the lesion's central point in the template scan and the detected points in follow-up scans. The predicted similarity scores are also shown. Follow-up 2 of group (b) are from CT volumes that do not include the lesion region of the template, so SAM matched unrelated points with a low score. Group (f) shows an example of using a lesion annotation from a follow-up scan as template to match lesions in earlier scans. It can help radiologists to check if the lesion has been missed in the earlier scan.}
	\label{fig:lesion_match} 
\end{figure*}

More qualitative examples on universal lesion matching are illustrated in \Fig{lesion_match}. SAM successfully matched a variety of lesions in different body parts, despite the appearance variance brought by image contrast, lesion size, field of view of the image, etc.

\bibliographystyle{IEEEtran}
\bibliography{tmi}

\end{document}